\documentclass[conference]{IEEEtran}
\IEEEoverridecommandlockouts
\usepackage{cite}
\usepackage{amsmath,amssymb,amsfonts}
\usepackage{algorithmic,algorithm}
\usepackage{graphicx}
\usepackage{textcomp}
\usepackage{xcolor}
\usepackage{subfigure}
\usepackage{booktabs}

\usepackage{etoolbox}
\makeatletter
\patchcmd{\@makecaption}
  {\scshape}
  {}
  {}
  {}
\makeatother

\def\BibTeX{{\rm B\kern-.05em{\sc i\kern-.025em b}\kern-.08em
    T\kern-.1667em\lower.7ex\hbox{E}\kern-.125emX}}
\begin{document}

\title{Efficient Distributed Framework for Collaborative Multi-Agent Reinforcement Learning\\
}

\author{\IEEEauthorblockN{1\textsuperscript{st} Shuhan Qi}
\IEEEauthorblockA{\textit{Computer Application Research Center} \\
\textit{Harbin Institute of Technology, Shenzhen}\\
Shenzhen, China \\
}
\and
\IEEEauthorblockN{2\textsuperscript{nd} Shuhao Zhang}
\IEEEauthorblockA{\textit{Computer Application Research Center} \\
\textit{Harbin Institute of Technology, Shenzhen}\\
Shenzhen, China} \\
\and
\IEEEauthorblockN{3\textsuperscript{rd} Xiaohan Hou}
\IEEEauthorblockA{\textit{Computer Application Research Center} \\
\textit{Harbin Institute of Technology, Shenzhen}\\
Shenzhen, China} \\
\and
\IEEEauthorblockN{4\textsuperscript{th} Jiajia Zhang}
\IEEEauthorblockA{\textit{Computer Application Research Center} \\
\textit{Harbin Institute of Technology, Shenzhen}\\
Shenzhen, China} \\
\and
\IEEEauthorblockN{5\textsuperscript{th} Xuan Wang}
\IEEEauthorblockA{\textit{Computer Application Research Center} \\
\textit{Harbin Institute of Technology, Shenzhen}\\
Shenzhen, China} \\
\and
\IEEEauthorblockN{6\textsuperscript{th} Jing Xiao}
\IEEEauthorblockA{\textit{Artificial Intelligence and Big Data Center} \\
\textit{Ping An Technology, Shenzhen}\\
Shenzhen, China} \\
}

\maketitle

\begin{abstract}
Multi-agent reinforcement learning for incomplete information environments has attracted extensive attention from researchers. However, due to the slow sample collection and poor sample exploration, there are still some problems in multi-agent reinforcement learning, such as unstable model iteration and low training efficiency. Moreover, most of the existing distributed framework are proposed for single-agent reinforcement learning and not suitable for multi-agent. In this paper, we design an distributed MARL framework based on the actor-work-learner architecture. In this framework,  multiple asynchronous environment interaction modules can be deployed simultaneously, which greatly improves the sample collection speed and sample diversity. Meanwhile, to make full use of computing resources, we decouple the model iteration from environment interaction, and thus accelerate the policy iteration. Finally, we verified the effectiveness of propose framework in MaCA military simulation environment and the SMAC 3D realtime strategy gaming environment with imcomplete information characteristics. 
\end{abstract}


\section{Introduction}
In general, reinforcement learning (RL) can be seen as a hybrid of game theory \cite{1} and dynamic strategy modeling \cite{2}.
Compared to traditional machine learning methods, RL performs policy update based on the feedback from the environment.
In this way, the RL can alter the policy dynamically and make more real-time decision in a changing environment.
Deep reinforcement learning technology \cite{3}, which combines reinforcement learning and deep learning, has been shown to have perceived decision-making abilities. 
In recent years, it has achieved spectacular results in the field of stock market trading \cite{4}, autonomous driving \cite{5}, robotics \cite{6}, machine translation \cite{7}, recommendation systems \cite{8}, and it is progressively expanding from many applications. 
\begin{figure}[h]
    \subfigure[$3m$ scenario]{
    \begin{minipage}[t]{0.5\linewidth}
    \includegraphics[width=4.3cm]{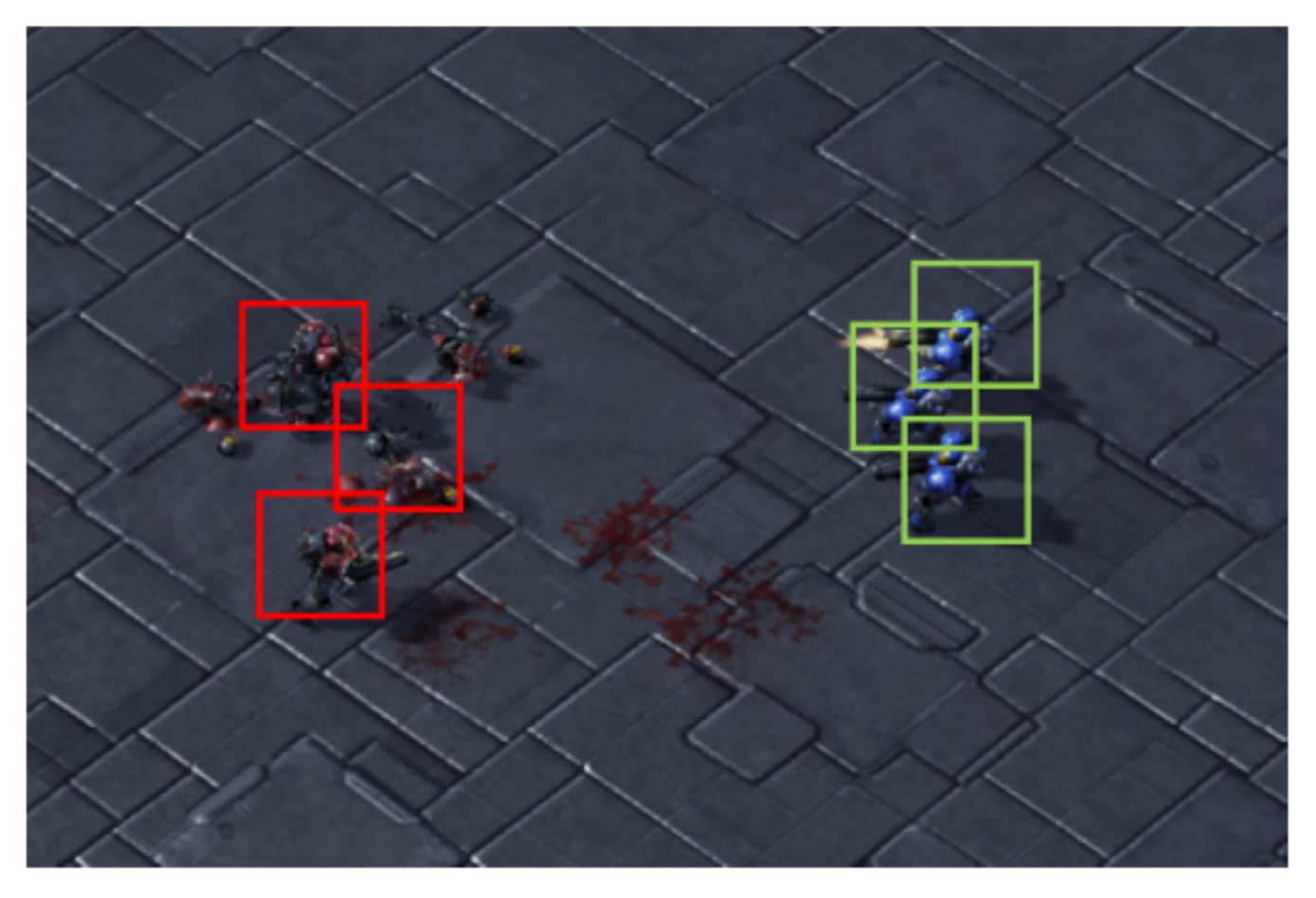}
    \label{subfig:11}
    \end{minipage}
    }%
    \subfigure[$8m$ scenario]{
    \begin{minipage}[t]{0.5\linewidth}
    \includegraphics[width=4.3cm]{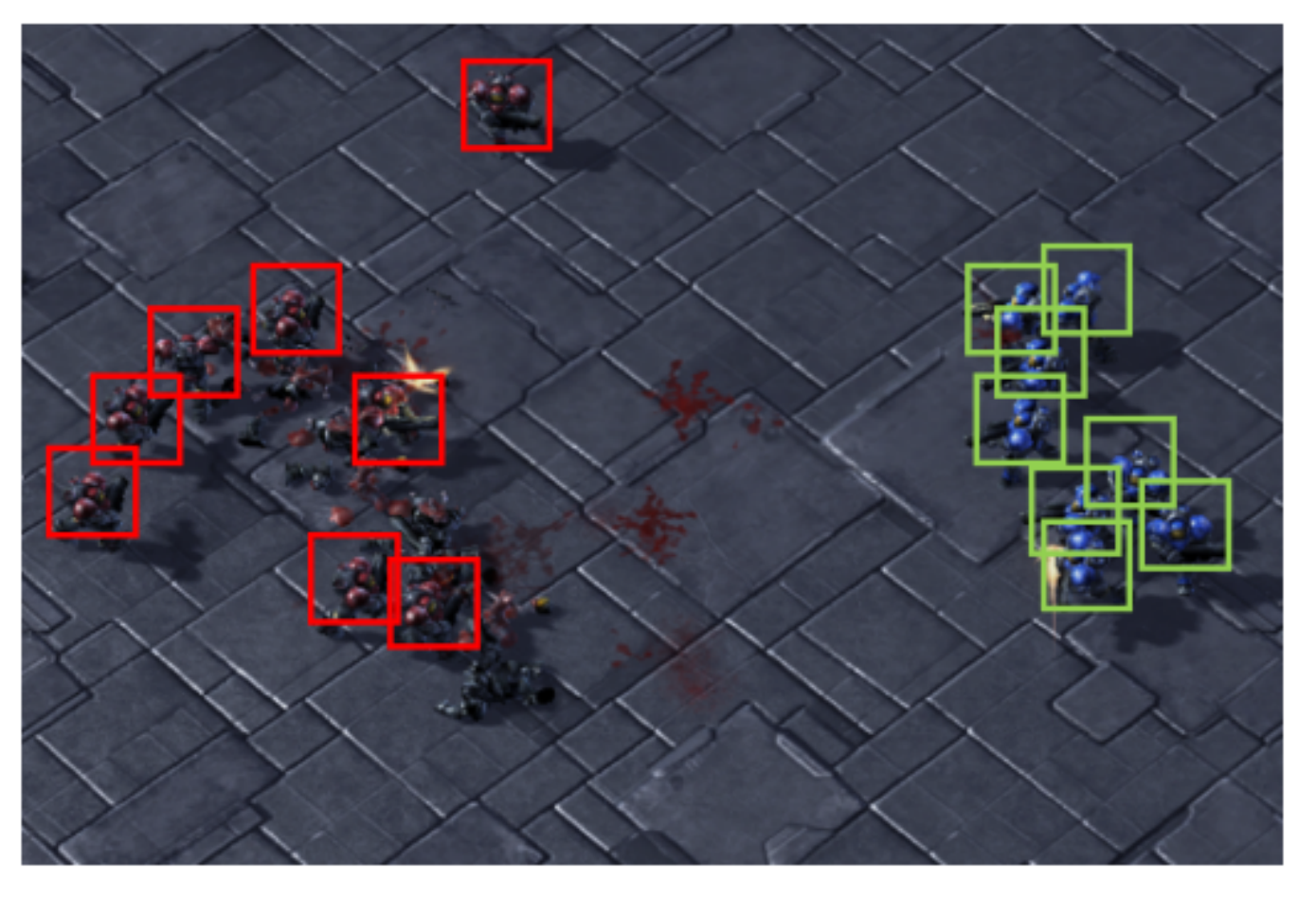}
    \label{subfig:12}
    \end{minipage}%
    }
    
    \subfigure[$2s3z$ scenario]{
    \begin{minipage}[t]{0.5\linewidth}
    \includegraphics[width=4.3cm]{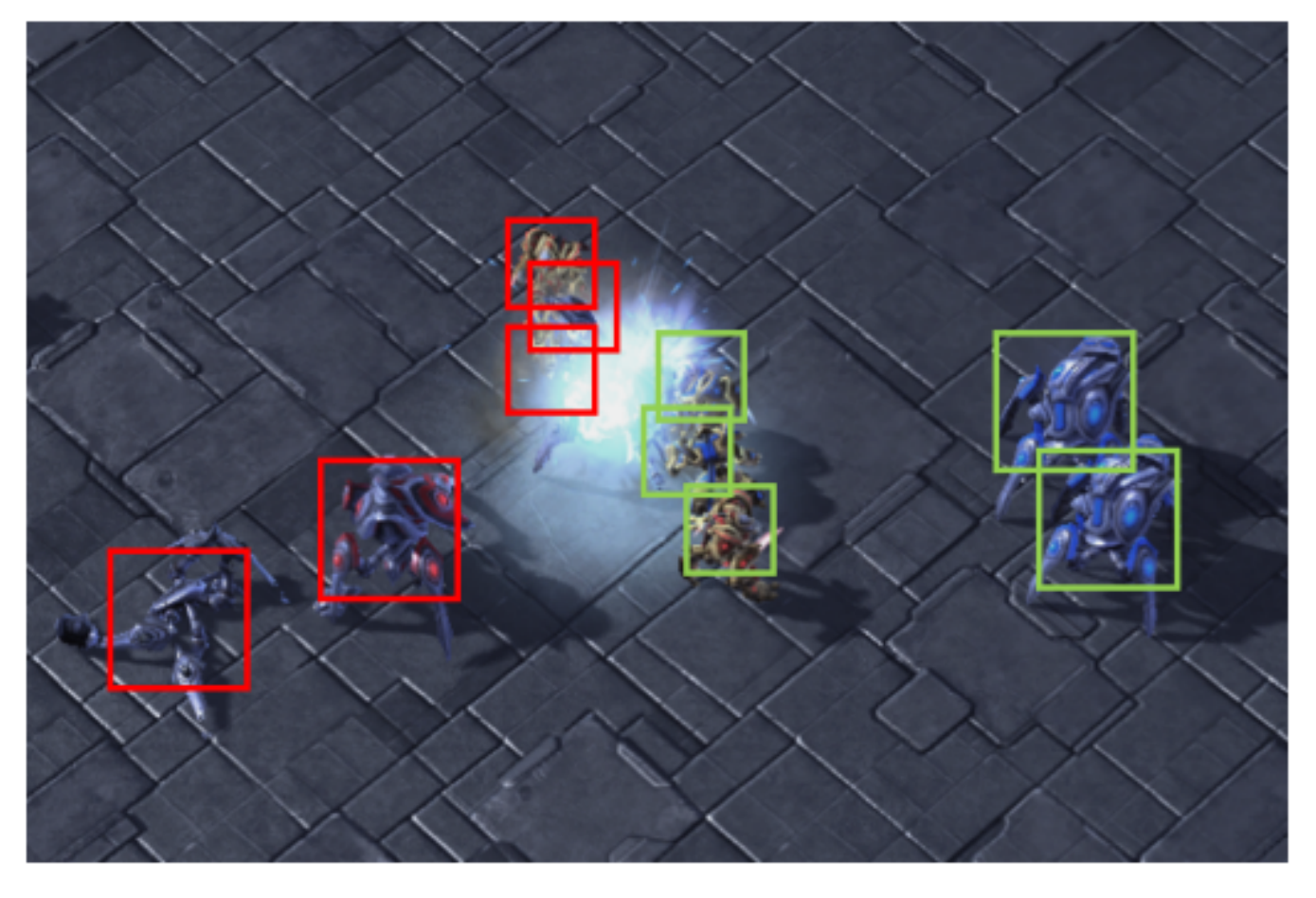}
    \label{subfig:13}
    \end{minipage}%
    }%
    \subfigure[$5 m_{-} v s_{-} 6 m$ scenario]{
    \begin{minipage}[t]{0.5\linewidth}
    \includegraphics[width=4.3cm]{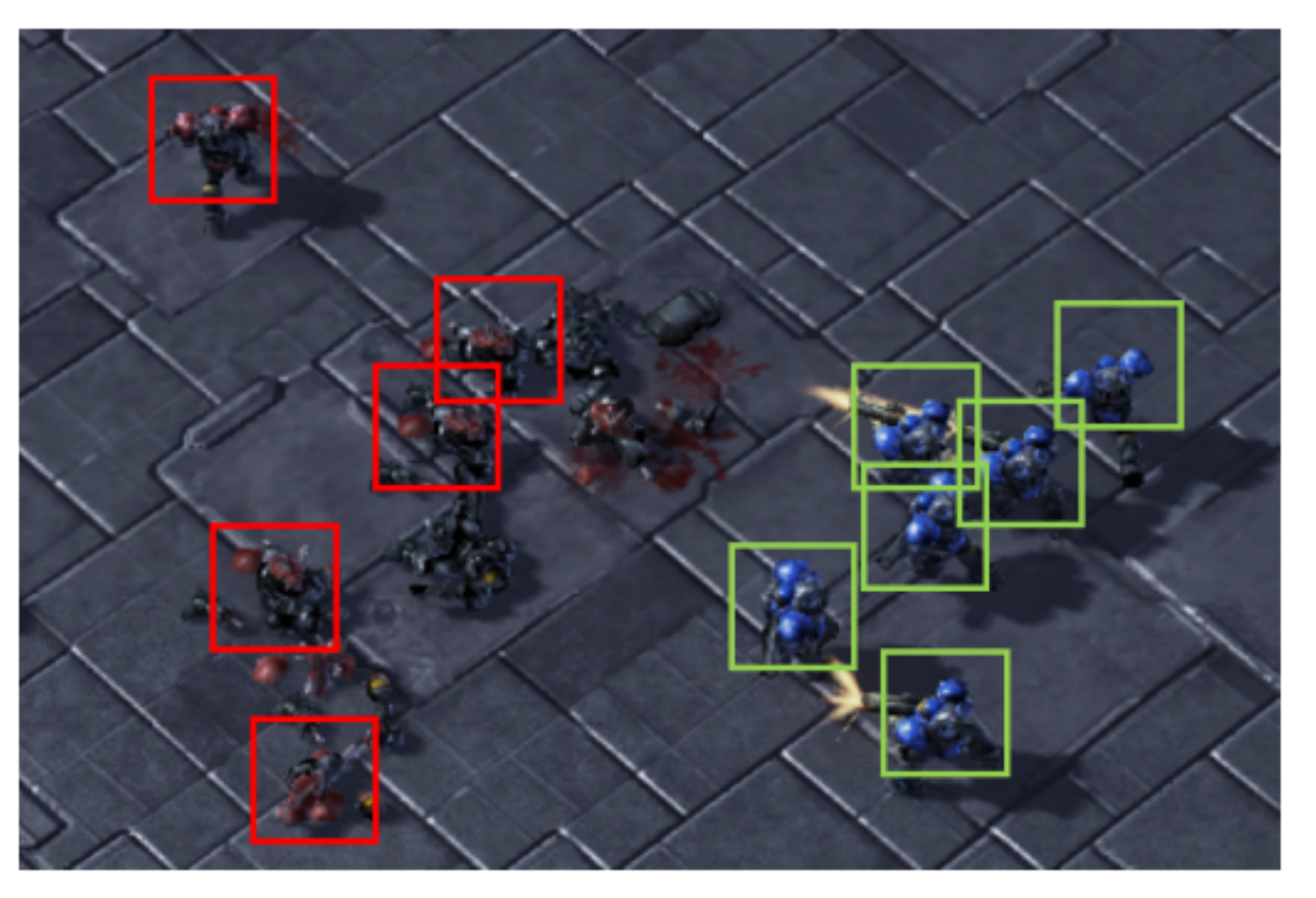}
    \label{subfig:14}
    \end{minipage}%
    }
\caption{The scenarios of experiment in SMAC.}
\label{fig:9}
\end{figure}

However, there are still many obstacles that limit the application of RL, such as lack of training samples, low sample utilization and unstable training, etc. Specifically, the training samples of RL are obtained through real-time interactions between the agent and the environment, which limits the diversity of training samples, and result to the situation of insufficient training samples. In particular, when the problem is extended to the field of multi-agent, the demand of training sample becomes even more severe. The scarce of sample will lead to the inefficiency of training and the instability of iteration for multi-agent model. What’s more, the effect of model training are also impacted by the diversity of empirical samples, which makes this problem even worse.
Therefore, increasing the amount of sample data can not only improve the training efficiency but also can reflect a more integrity of the environment, and thus resulting in a stable agent learning.

Researchers have lately begun to investigate distributed reinforcement learning (distributed RL) domains. The distributed framework can be used to boost data throughput and solve the problems of samples scarce and unstable training, which is critical to the reinforcement learning training process. However, the most of the current distributed RL framework is proposed for single-agent RL, and not suitable for the solving the  multi-agent problem.

To address the above problems, in this paper,  we proposed two practical distributed training frameworks for MARL.
Firstly, we propose a actor-worker based MARL training framework. The framework consists of a worker and many actors, and information is transmitted between worker and actors through pipelines. It worth noting that, in this framework, the actors interact with multiple environments in an asynchronous way, which significantly improve the efficiency of sample collection and thus reduce the cost of multi-agent training.

Moreover, we extend the actor-work to  actor-worker-learner distributed framework (DMCAF), which further decouples environment interaction from the model iteration process.
Unlike the A3C method that transfers gradients, this framework merely transmits observation data from actors to workers so that actors do not need to participate the model training.
This further improves ample diversity and policy iteration efficiency, and makes the multi-agent training process becomes more stable and fast. 
\begin{figure*}[t]
    \subfigure[On-policy]{
    \begin{minipage}[t]{0.28\linewidth}
    \includegraphics[width=5cm]{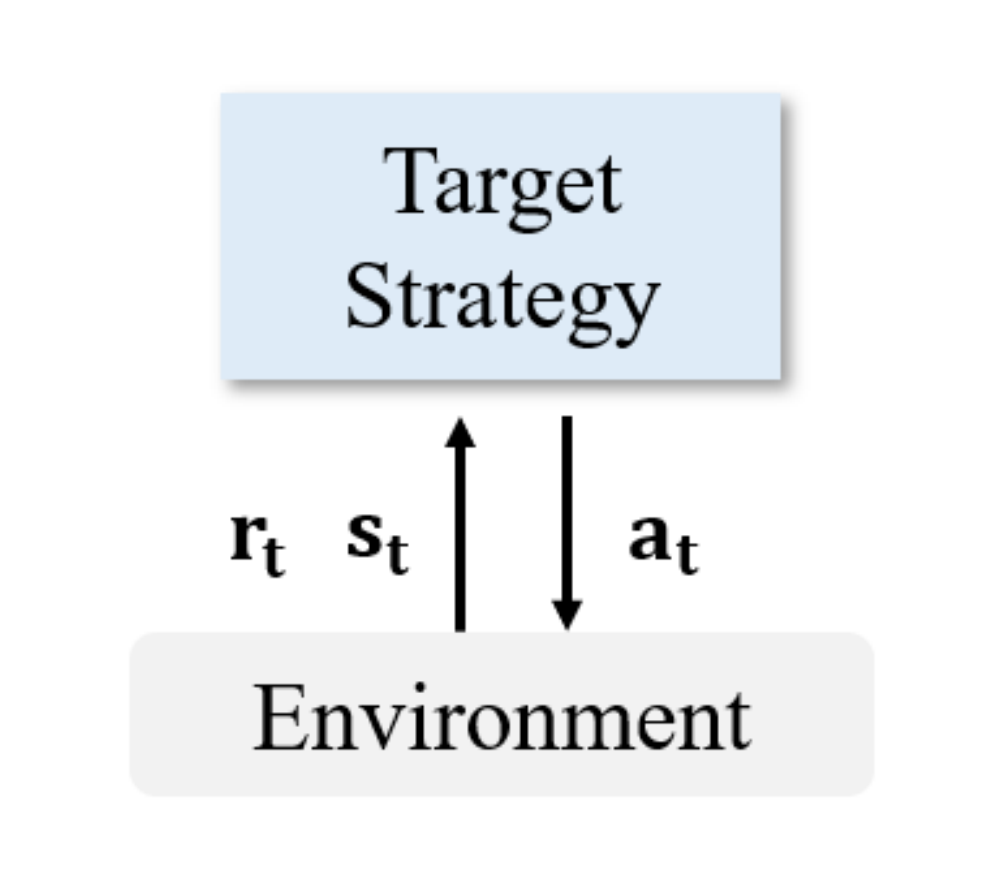}
    \label{subfig:1}
    \end{minipage}%
    }%
    \subfigure[Off-policy]{
    \begin{minipage}[t]{0.72\linewidth}
    \includegraphics[width=13cm]{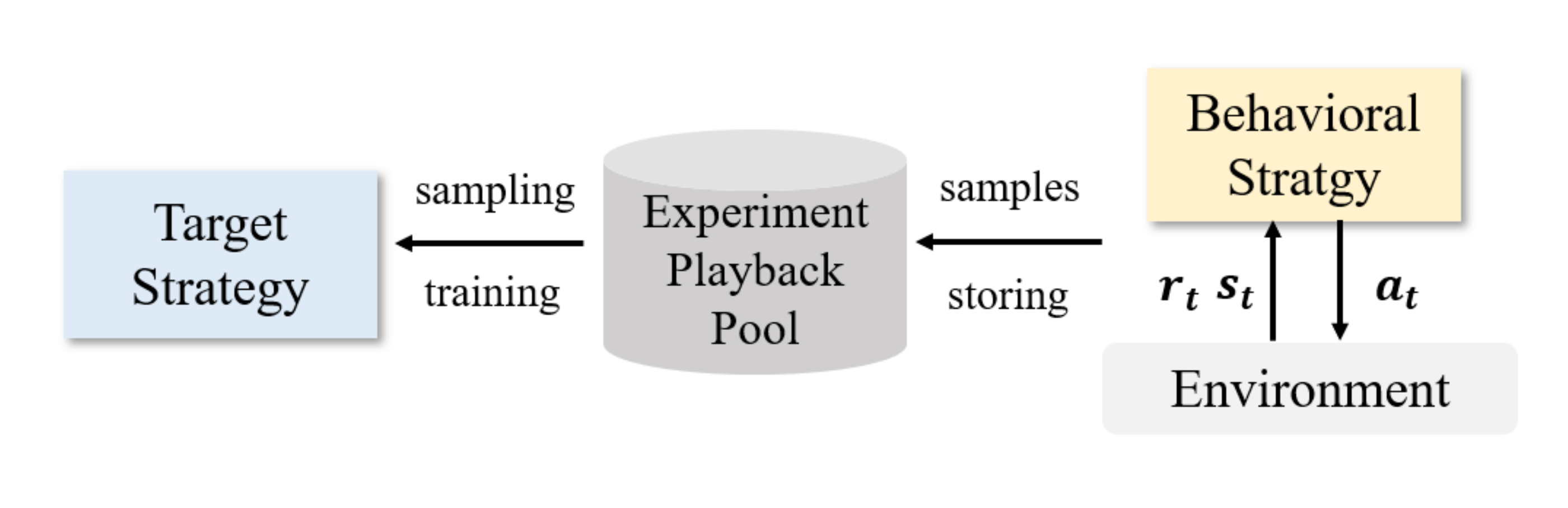}
    \label{subfig:2}
    \end{minipage}%
    }%
\caption{On-policy and off-policy}
\label{fig:1}
\end{figure*}

\section{Background}
\subsection{Related Work in Reinforcement Learning}
Watkins created a table-based Q-learning reinforcement learning approach in 1989, which laid the groundwork for today's advancement of RL. 
But it has the issue that the table needed may be large in case of numerous states.
The DeepMind team proposed the DQN (Deep Q Network) method in 2013 to solve the issue of Q-learning. 
The DQN technique trains Atari game bots to outperform human players by combining the target network and experience replay mechanism \cite{10}. 
Following its triumph in conquering the Atari game, Google DeepMind's Go agent AlphaGo has now defeated top Go masters Li Sedol and Ke Jie \cite{11}. 
The enhanced version AlphaGoZero surpasses the human level in just a few days through self-play \cite{12,13}.
In addition to Go, DouZero \cite{14}, a chess and card-type landlord agent developed by the domestic Kuaishou team, and Suphx \cite{15}, a mahjong agent developed by the Microsoft team, combine RL, supervised learning, self-play, and monte carlo search that have made breakthroughs in different chess and card fields. 
In real-time strategy games with larger dimensions and more complex environments, the agent AlphaStar \cite{16} developed by DeepMind shows an excellent game level on StarCraft II.




In the field of distributed RL, the actor-critic algorithm is a classic method, which combines policy and value networks, greatly improving the level of the agent. 
The A3C algorithm is an asynchronous distributed version of the actor-critic algorithm in single-agent RL \cite{17}, which maintains multiple secondary networks and a primary network. 
Among them, each secondary network contains actors $\theta^{\prime}$ and critics $\omega^{\prime}$ and the primary network contains actors $\theta$ and critics $\omega$, which have the same network structure. 
After interacting with the environment, the secondary network computes and accumulates the  gradients of actor and critic:
\begin{equation}
    \centering
    d \theta \leftarrow d \theta+\nabla_{\theta^{\prime}} \log \pi_{\theta^{\prime}}(a \mid s)\left(R-V_{w^{\prime}}(s)\right)
    \label{equ:3}
\end{equation}
\begin{equation}
    \centering
    d w \leftarrow d w+\partial\left(R-V_{w^{\prime}}(s)\right) / \partial w^{\prime}
    \label{equ:4}
\end{equation}

The primary network receives the gradients accumulated by the secondary networks, and updates the gradients for actor $\theta$ and critic $\omega$ with learning rates $\alpha$ and $\beta$, respectively:
\begin{equation}
    \centering
    \theta \leftarrow \theta - \alpha d \theta
    \label{equ:5}
\end{equation}
\begin{equation}
    \centering
    \omega \leftarrow \omega - \beta d \omega
    \label{equ:6}
\end{equation}

The GA3C method proposed by Babaeizadeh et al. uses multiple actors to interact with the environment. 
Each actor only transmits the environment state and training samples to the learner through a queue, which reduces the number of policy models and training resources \cite{18}. 
The IMPALA algorithm proposed by Deepmind improves A3C from another dimension. 
Different from the gradient transfer in A3C, IMPALA adopts the method of transferring training samples to update the model and proposes the V-trace method to correct the policy-lagging issue\cite{19}. 
The Ape-X algorithm proposed by Horgan et al. is a value-based distributed RL method. 
This method uses multiple processes to realize multiple environment interactions, which improves the training speed of the single-process DQN \cite{20}.

\subsection{On-policy and Off-policy}
According to the policy which generates training data, distributed MARL can be divided into on-poliy and off-policy.
As shown in Fig.~\ref{subfig:1}, if the experience samples sampled by the agent during training come from the current target strategy, this method is called on-policy learning \cite{22}. 
Off-policy learning \cite{23} is the method that employs an exploratory behavioral strategy to interact with the environment and gather experience.
And it also stores experience samples in the experience replay pool \cite{24} for sampling and iterative updating of the target strategy, which is shown in Fig.~\ref{subfig:2}. 

One drawback of the on-policy learning is the need to guarantee that the parameters of the model network in each environmental process are consistent with the parameters of the primary model network, as well as the synchronization of each environmental interaction. 
Thus when two environments are out of sync, they must wait for each other. 
Another drawback is the ability to allow asynchronous sampling of the environment, which solves the problem of low training efficiency but brings the additional policy-lagging issue.
The training samples collected by the primary model network are policy-lagging, meaning they are obtained based on the past policy rather than the current policy. 
So during the iterative update of the policy, the primary model network must execute importance sampling on the experience samples, turning them into learnable experience samples.

\begin{footnotesize}
    \begin{equation}
    \centering
        \begin{aligned}
        \mathbb{E}_{x \sim p(x)}[f(x)] &=\int f(x) p(x) d x \\
        &=\int \frac{q(x)}{q(x)} f(x) p(x) d x=\int q(x) \frac{p(x)}{q(x)} f(x) d x\\
        \end{aligned}
    \label{equ:1}
    \end{equation}
\end{footnotesize}
\begin{footnotesize}
    \begin{equation}
    \hspace{-7.5mm}
    =\mathbb{E}_{x \sim q(x)}\left[\frac{p(x)}{q(x)} f(x)\right]
    \label{equ:2}
    \end{equation}
\end{footnotesize}

The importance sampling is created to handle the problem of calculating the expectation using a known distribution $q(x)$ When the original distribution $p(x)$ is difficult to sample, which is shown in \eqref{equ:1} and \eqref{equ:2}.

In on-policy distributed RL, the importance sampling is used to remedy the policy-lagging problem produced by samples from different policies. 
But the off-policy distributed RL does not require as many modifications to the samples and expectation computations due to the presence of the experience replay mechanism. 
Each environmental process can store the collected samples into the experience replay pool at any time, and the primary network takes out the historical sample data from the experience replay pool to iteratively update the model.

\subsection{The Meaning of Exploring distributed MARL}
With the development of MARL, RL has been able to deal with the common incomplete information and multi-agent challenges in the environment \cite{25}.
But the instability of training and performance have become the factors restricting its development.
And from the results of distributed single-agent reinforcement learning algorithms, distributed technology is useful for improving performance.

However, the methods in distributed RL are nearly in the single-agent domain, and it is difficult to directly extend them to the multi-agent domain.
Specifically, the model calculation becomes more complicated as the number of agents increased, so does the sample requirement for multi-agent model training. 

As a result, the research on distributed MARL is of great significance for improving the stability and performance.
Therefore,we propose the DMCAF in this paper.



\section{Method}
For the reasons stated above, we focus on proposing a distributed MARL method. 
To avoid the high-complexity problem caused by policy correction technology \cite{30}, we select the value-based MACRL method as the basic algorithmic model and propose a distributed MARL algorithmic framework.
The algorithmic framework we proposed is mainly divided into two parts, the latter part is the continuous deepening and extension of the former part.

\subsection{Actor-worker Distributed Multi-agent Asynchronous Communication Framework}\label{AA}
\begin{figure}[htbp]
    \centering
    \includegraphics[width=9cm]{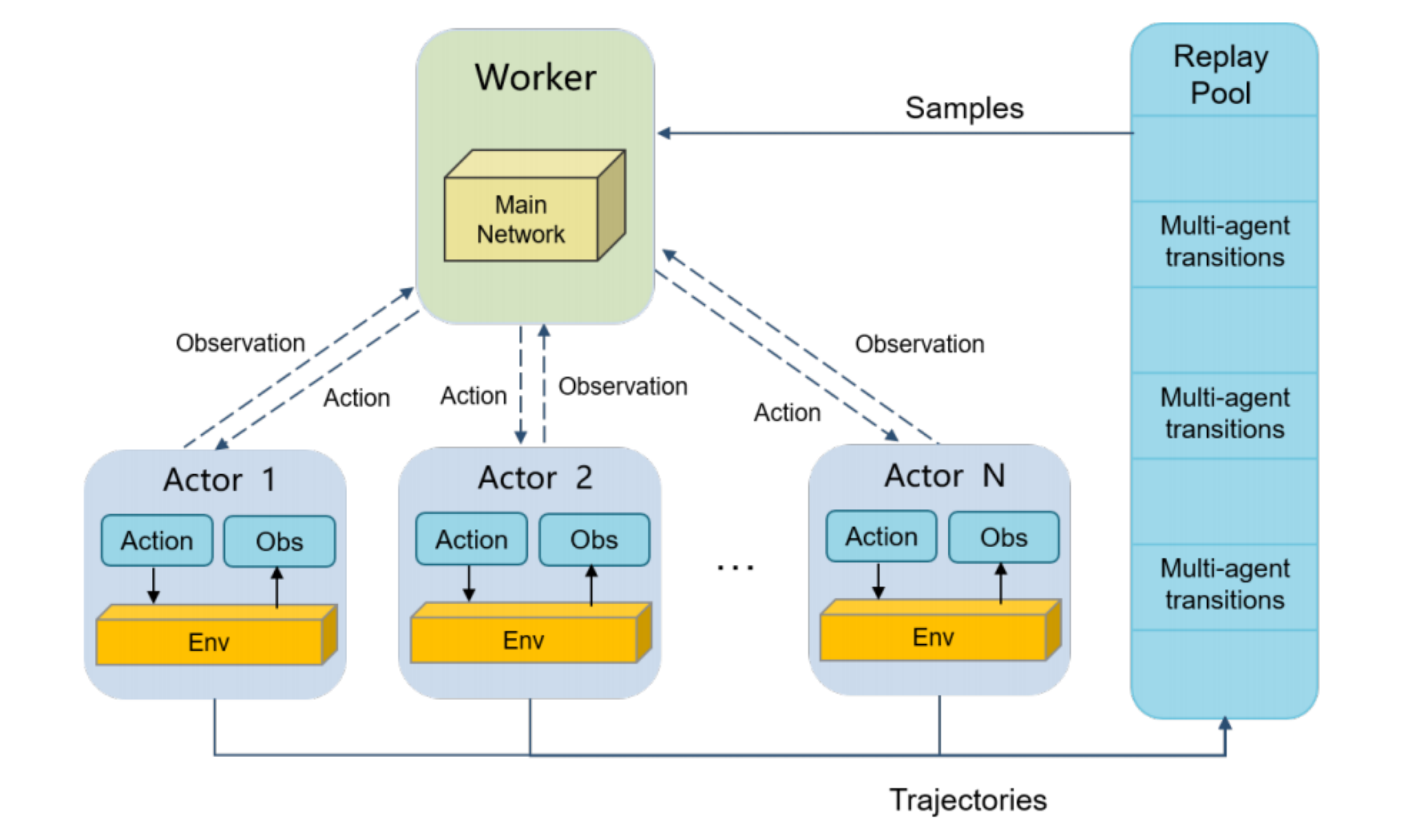}
    \caption{Actor-worker asynchronous communication framework.The framework is made up of a worker and numerous actors. Each actor in the multi-process maintains an environment and communicates asynchronously with the worker via pipeline. There exists an observation-action pipeline and an experience pipeline between each actor and worker.}
    \label{fig:2}
\end{figure}


We first propose an actor-worker distributed algorithm training framework for multiple agents.
The algorithmic training framework is shown in Fig.~\ref{fig:2}.
The operation of the framework is divided into four steps.
\begin{itemize}
    \item Step 1: The environment passes the new state through the observation-action pipeline in actor to the worker at each time step.
    \item Step 2: The worker sends the state into its primary network for decision-making and puts the action back into the corresponding observation-action pipeline which is passed to the corresponding actor.
    \item Step 3: The environment gets the action from pipeline and goes to the next state.
    \item Step 4: After each round of data collection, the actor sends the entire gathered trajectory to the experience replay pool via the experience pipeline.
    The worker randomly samples from the updated experience replay pool to update the policy until it iteratively converges.
\end{itemize}

\begin{figure}[htbp]
    \centering
    \includegraphics[width=9cm]{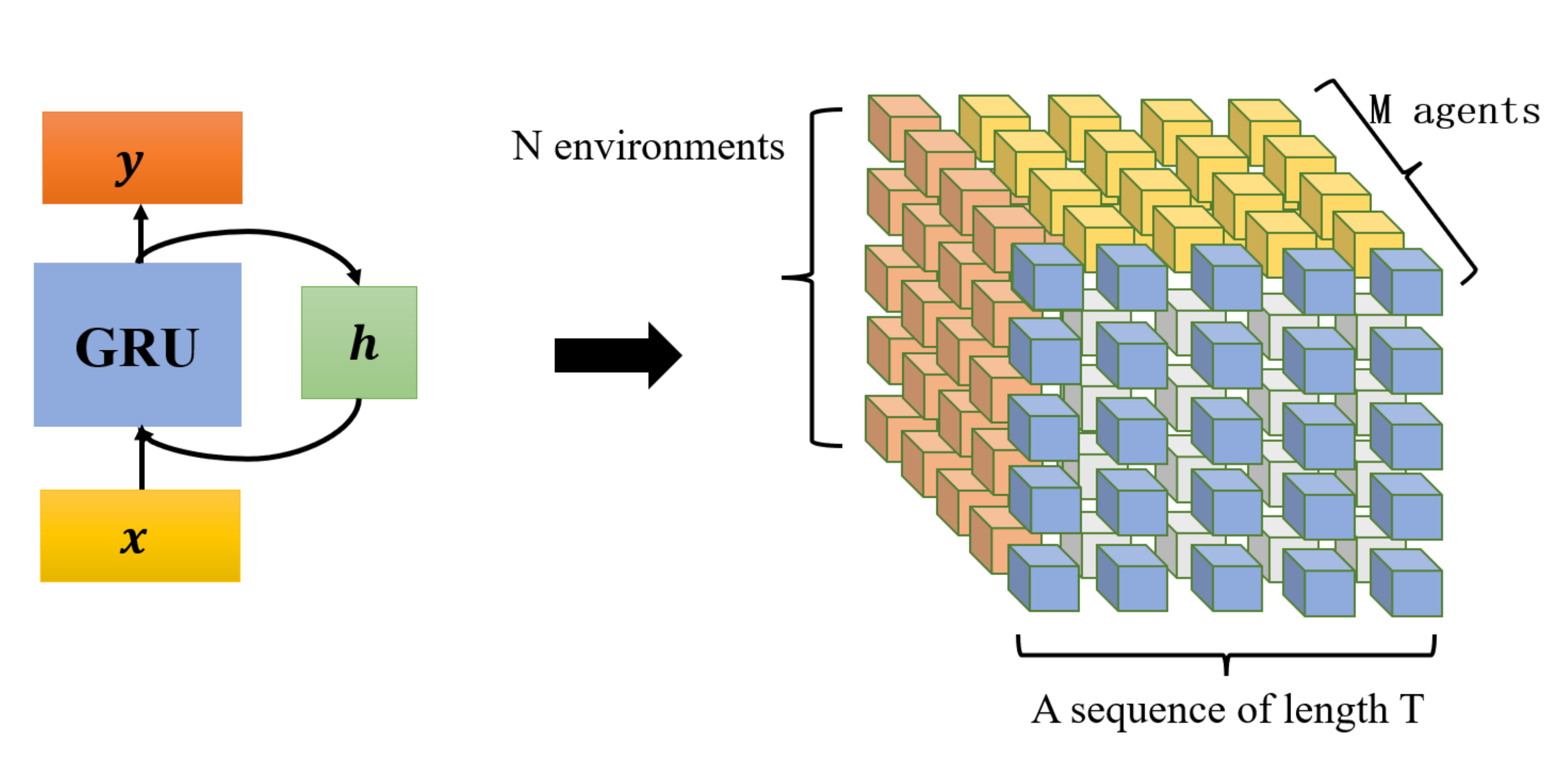}
    \caption{Multidimensional hidden state. When M agents make a sequence choice of length T in N environments, the space of hidden state is expanded from single to multi-dimensional, with each dimension corresponding to the number of agents, environments, and time steps, preserving all hidden state spaces.}
    \label{fig:3}
\end{figure}

However,when employing the training framework described above, the historical knowledge shared by multiple agents in different contexts must be explicitly split. 
The hidden state unit in the distributed MARL method is shown in Fig.~\ref{fig:3}. 
Otherwise, the agents in different environments will misuse historical information, lowering the performance of the algorithm.

After the foregoing modifications, multiple environments can be deployed simultaneously. Therefore, the speed and scalability of sample collecting will be increased. And moreover, the asynchronous interaction of multiple actors further lowers the correlation between samples. 
Meanwhile, because all policy updates are accomplished iteratively in workers, each actor just need to maintains a distinct environment instead of the model with large amount of parameters, thus lowering the computational cost of large-scale model training.

\subsection{DMCAF:Actor-worker-learner Distributed Multi-agent Decoupling Algorithmic Framework}
The worker-learner layer is presented after proposing the actor-worker framework, and model sharing of the primary network and the secondary network is implemented in the worker-learner layer. 
The actor-worker structure is extended to the actor-worker-learner structure.

The A3C algorithm is the most representative asynchronous algorithm, but it has two problems.
The first is that the primary network is updated slowly and has poor robustness. 
Secondly, this way of passing gradients makes the model iteratively slower.
So we use the method of transferring empirical samples from the secondary network to the primary network instead of transferring gradients. 
In this way, the data collection and model updates can be decoupled. 
This decoupling approach can well connect the actor-worker asynchronous communication framework proposed in the previous section.
Therefore, we take the actor-worker asynchronous communication framework as the underlying structure, and introduce the decoupled worker-learner structure as the model sharing layer.

\begin{figure}[htbp]
    \centering
    \includegraphics[width=9cm]{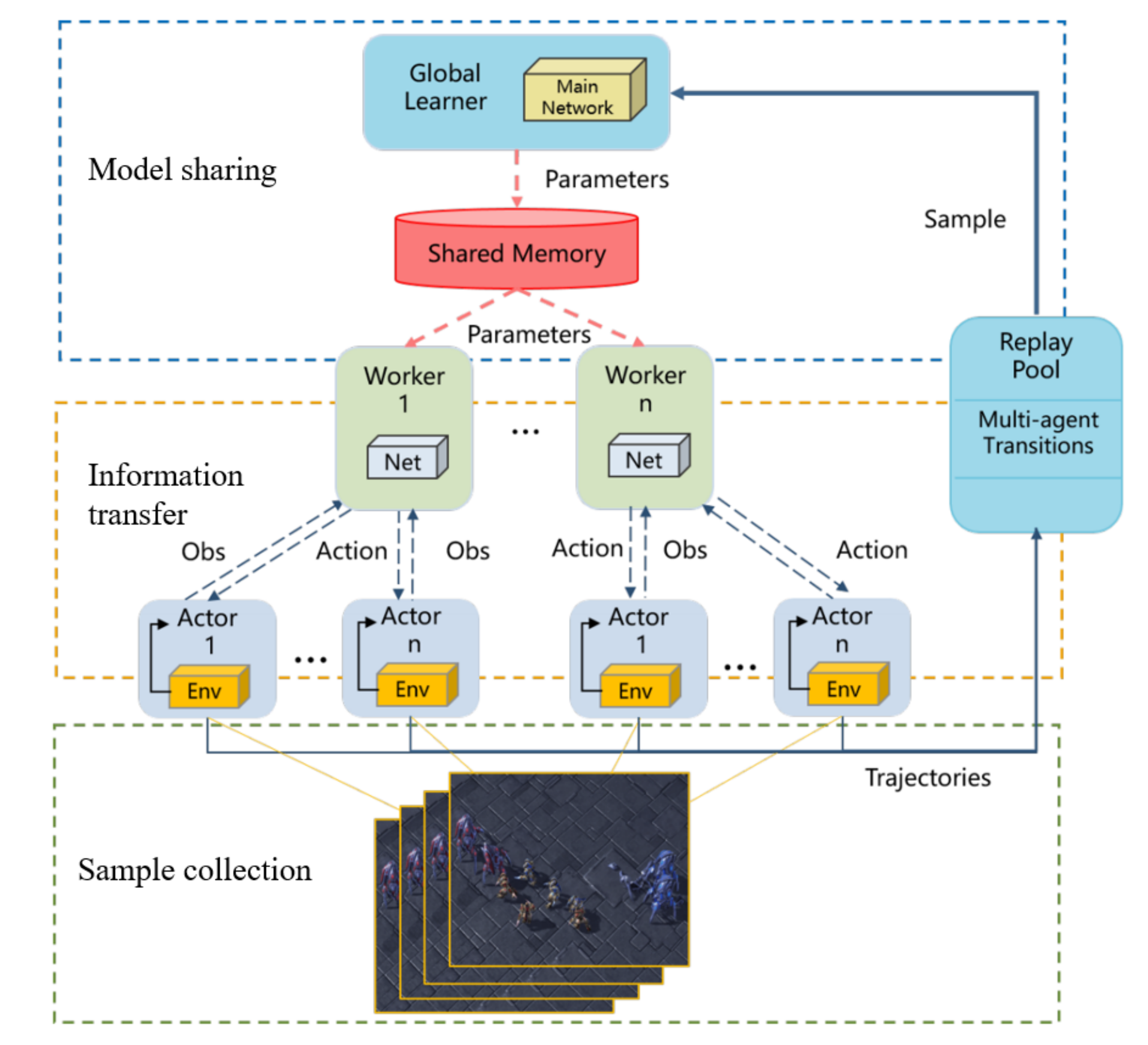}
    \caption{Actor-worker-learner distributed decoupling framework.The three-layer structure corresponds to the actors, workers and learners. Learners, workers, and actors cooperate with each other from top to bottom to complete the functions of the model sharing, information transmission and sample collection.}
    \label{fig:4}
\end{figure}

Specifically, under this decoupling framework, although the worker has a policy network, it is only responsible for decision-making and sample collection during training.
The worker does not calculate and transmit gradients, thus avoiding the problem of gradient conflict. 
After multiple actors collect the experience samples of the whole game through interaction, the workers send the samples to the experience replay pool, and the learner takes experience from the experience replay pool to update the policy as needed.
The samples in the experience replay pool can be updated continuously because the asynchronous interaction of multiple environments maintained by the actor greatly improves the speed of data collection.
So the learner does not need to wait for the gradient to be calculated or for data to be collected.

In order to promote further optimization of training, in terms of the model sharing, we use the shared memory technology to periodically update the parameters of the primary network in learner to the secondary network in worker instead of transferring parameters. 
The use of shared memory technology enables the proposed method to avoid the instability of data transmission when synchronizing the model, and reduces the communication cost while improving the communication efficiency.

In summary, we propose the actor-worker-learner distributed decoupling framework, as shown in Fig.~\ref{fig:4}. 
The worker layer acts as an intermediate transition layer, maintaining multiple actors and a neural network that periodically synchronizes with the learner. 
The worker starts training with multiple actors at the same time and communicates with each actor. 
Each actor maintains an independent environment, and each of them interacts with the environment under the guidance of the policy network in the worker to complete the sample collection. 
The algorithm of the actor is shown in Algorithm \ref{alg:1}.

\subsubsection{Actor Algorithm}
When the actor receives the new state of the environment, it delivers the observation into the observation-action pipeline, and the action of the worker is retrieved from the observation-action pipeline to interact with the environment. 
Each step of interactions between the agent and the environment is recorded by the actor, and at the end of the game, the entire game experience is transmitted to the sample queue for the learner to use.
\begin{algorithm}[t]
\caption{Actor Algorithm} 
\textbf{Input:} environment parameters $args$,
                              interaction times $T$\\ 
\textbf{Global:} sample queue $Q$,
                            observation-action pipe $P$
                         
\begin{algorithmic}[1]
\STATE \textbf{Function} Actor($args$, $T$)\textbf{:}
    \STATE Construct environment $Env$ from environment           parameters $args$;
    \STATE $Env.start()$;\\
    \STATE Initialize interaction times $t \leftarrow 0$;\\
    \REPEAT
        \STATE $Env.reset()$;
        \STATE Get the start state;
        \REPEAT
            \STATE Put the observation $o$ into observation-action pipe $P$;
            \STATE Get action $a$ from observation-action pipeline $P$;
            \STATE $o,r \leftarrow Env.step(a)$;
        \UNTIL{$Env.done()$};
        \STATE $t \leftarrow t+1$;
        \STATE Send the collected whole track to the sample queue $Q$;
    \UNTIL{$t > T$};
    \STATE $Env.close()$;
\RETURN
\end{algorithmic}
\label{alg:1}
\end{algorithm}

\subsubsection{Worker Algorithm}
The decoupled workers and learners work asynchronously while training. 
The algorithm of the worker is shown in \ref{alg:2}. 
The worker deploys multiple actors, gets observations from the observation-action pipeline, makes decisions using the policy network that are periodically synchronized from the shared memory pool.
Then it pips actions back to the actors. 
When the worker fails to get new observations from the observation-action pipeline, the loop ends.
\begin{algorithm}[h]
\caption{Worker Algorithm} 
\textbf{Input:} the number of actors $n$,
                              total interaction times $nT$\\ 
\textbf{Global:} network parameters $\theta$,
                            shared memory pool $M$,
                            observation-action pipeline $P$
                         
\begin{algorithmic}[1]
\STATE \textbf{Function} Worker($n$, $nT$)\textbf{:}
    \STATE Deploy $n$ actors and perform $T$ environment interactions respectively;
    \REPEAT
        \STATE Get parameters $\theta$ from shared memory pool $M$ or policy synchronization;
        \STATE Get observations $o$ from observation-action pipeline $P$;
        \STATE Make decisions based on  policy,$a \sim \pi(a \mid o)$;
        \STATE Send action $a$ into the observation-action pipeline; $P$;
    \UNTIL{$P.size == 0$};
\RETURN
\end{algorithmic}
\label{alg:2}
\end{algorithm}

\subsubsection{Learner Algorithm}
The learner maintains the sample queue, shared memory pool, and experience replay pool. 
The cyclic iterative training of model begins when multiple workers are initiated. 
The sample queue sends a huge number of samples created by the interaction of different environments to the shared experience replay pool. 
When the experience pool is not empty, the learner continues to train and update the model.
Meanwhile, it synchronizes the most recent model parameters to the shared memory pool, allowing workers to interact with the environment using the latest policy model to gain fresh experience samples.
The training comes to a halt when the sample queue is empty, indicating that no environment is interacting.
\begin{algorithm}
\caption{Learner Algorithm} 
\textbf{Input:} the number of workers $m$,
                              batch size $B$\\ 
\textbf{Global:} experience replay pool $E$,
                            shared memory pool $M$,
                            sample queue $Q$,
                            network parameters $\theta$
                         
\begin{algorithmic}[1]
\STATE \textbf{Function} Learner($m$, $B$)\textbf{:}
    \STATE Initialize network $\theta,E,Q,M$;
    \STATE Deploy $n$ workers;
    \REPEAT
        \STATE Get data from the sample queue $Q$ and store it in the experience replay pool $E$;        
        \STATE Randomly sample $B$ samples from the experience replay pool $E$;
        \STATE Calculate the gradient:\\
        $d \theta \leftarrow\left(y_{j}^{\text {total }}-Q_{\text {total }}(s, \boldsymbol{a} ; \theta)\right) \nabla_{\theta} Q_{\text {total }}(s, \boldsymbol{a} ; \theta)$;
        \STATE Update parameter $\theta$:\\
        $\theta \leftarrow \theta+d \theta$;
        \STATE Sync model parameters $\theta$ to shared memory pool $M$;
    \UNTIL{$Q.size == 0$};
\RETURN
\end{algorithmic}
\label{alg:3}
\end{algorithm}

\begin{table*}[t]
    \centering
    \begin{tabular}{c|c|c|c|c|c}
        \toprule
         Scenario Name &  Number of Actions & Number of Agents & State Dimension & Observation Dimension & Time-step limit \\
        \hline
        $3m$ & 9 & 3 & 48 & 30 & 60 \\
        $8m$ & 14 & 8 & 168 & 80 & 120\\
        $2s3z$ & 11 & 5 & 120 & 80 &120\\
        $5 m_{-} v s_{-} 6 m$ & 12 & 5 & 98 & 55 & 70\\
        \bottomrule
    \end{tabular}
    \caption{The feature information of the agent in each scenario}
    \label{tab:2}
\end{table*}

\begin{table}[h]
    \centering
    \begin{tabular}{c|c|c|c}
        \toprule
         Scenario Name &  Own Unit & Enemy Unit & scenario Type \\
        \hline
        $3m$ & $3 M$ & $3 M$ & $IS$\\
        $8m$ & $8 M$ & $8 M$ & $IS$\\
        $2s3z$ & $2 S$ and $3 Z$ & $2 S$ and $3 Z$ & $HS$\\
        $5 m_{-} v s_{-} 6 m$ & $5 M$ & $6 M$ & $IA$ \\
        \bottomrule
    \end{tabular}
    \caption{Scenario properties.We use $IS$ for isomorphic symmetry, $HS$ for heterogeneous symmetry, $IA$ for isomorphic asymmetric, $M$ for Marines, $S$ for Stalker, $Z$ for Zealots.}
    \label{tab:1}
\end{table}

This framework has strong scalability and can configure several actors and workers in a flexible manner. 
The upper-layer decoupling framework separates the data acquisition and model training modules, allowing for higher data throughput, more efficient model iteration, and better GPU use. 
The deployment of large-scale environments has substantially improved the sample collecting speed, extending the benefits of distributed systems even further. 
The asynchronous updating of numerous models promotes agent exploration in terms of policies and alleviates the exploration and utilization difficulties in RL.
Meanwhile, the correlation of RL training samples is weakened, and the diversity of samples is increased, thereby improving the robustness of the algorithm and making the training process more stable.

\section{Experiment}
\subsection{Environment}
SMAC (StarCraft Multi-Agent Challenge) \cite{26} is a MARL research platform about the game StarCraft II. 
It is designed to evaluate the ability of multi-agent models to solve complex tasks through collaboration. 
Many researchers have regarded the SMAC platform as a benchmark platform in the field of MACRL.

The difficulty on the SMAC platform can be customized, and we use the "extremely challenging" mode with a level of $7$ in the experiment. 
On the SMAC platform, each scenario depicts a battle between two forces. 
Depending on whether the type is the same or not, battle scenarios can be separated into symmetrical and asymmetric battle scenarios, as well as isomorphic and heterogeneous battle scenarios based on the unit types on both sides.

The experimental environment is comprised of four scenarios: $3m$, $8m$, $2s3z$, and $5 m_{-} v s_{-} 6 m$,which is shown in Fig.~\ref{fig:9}. 
The $3m$ isomorphic symmetric scenario, $8m$ isomorphic symmetric scenario, and $2s3z$ heterogeneous symmetric scenario are among the most commonly used experimental scenarios for MACRL.
And they are used in the experiments of the original papers of the Qmix and CoMA algorithms. 
The necessary characteristic information for the four cases is listed in Table \ref{tab:1}.

We use the control variable method to uniformly set the observations, states, actions, and rewards of all scenarios to the default values of the platform. 
The specific settings are listed in Table \ref{tab:2}.

\subsection{Experimental Display and Results Discussion}
We select the MACRL algorithm that is currently optimal in multiple scenarios of the SMAC platform as the baseline algorithm. 
Under the same scenario settings, the distributed multi-agent collaborative algorithm proposed is verified from the aspects of the speed of sample collection and performance.
\begin{itemize}
    \item \textbf{Qmix:}The original Qmix algorithm, which currently achieves optimal performance in multiple scenarios on the SMAC platform.
    \item \textbf{AW-Qmix:}Distributed Qmix algorithm based on Actor-Worker framework.
    \item \textbf{AWL-Qmix:}Distributed Qmix algorithm based on DMCAF.
\end{itemize}

\begin{figure}[htbp]
    \centering
    \includegraphics[width=8cm]{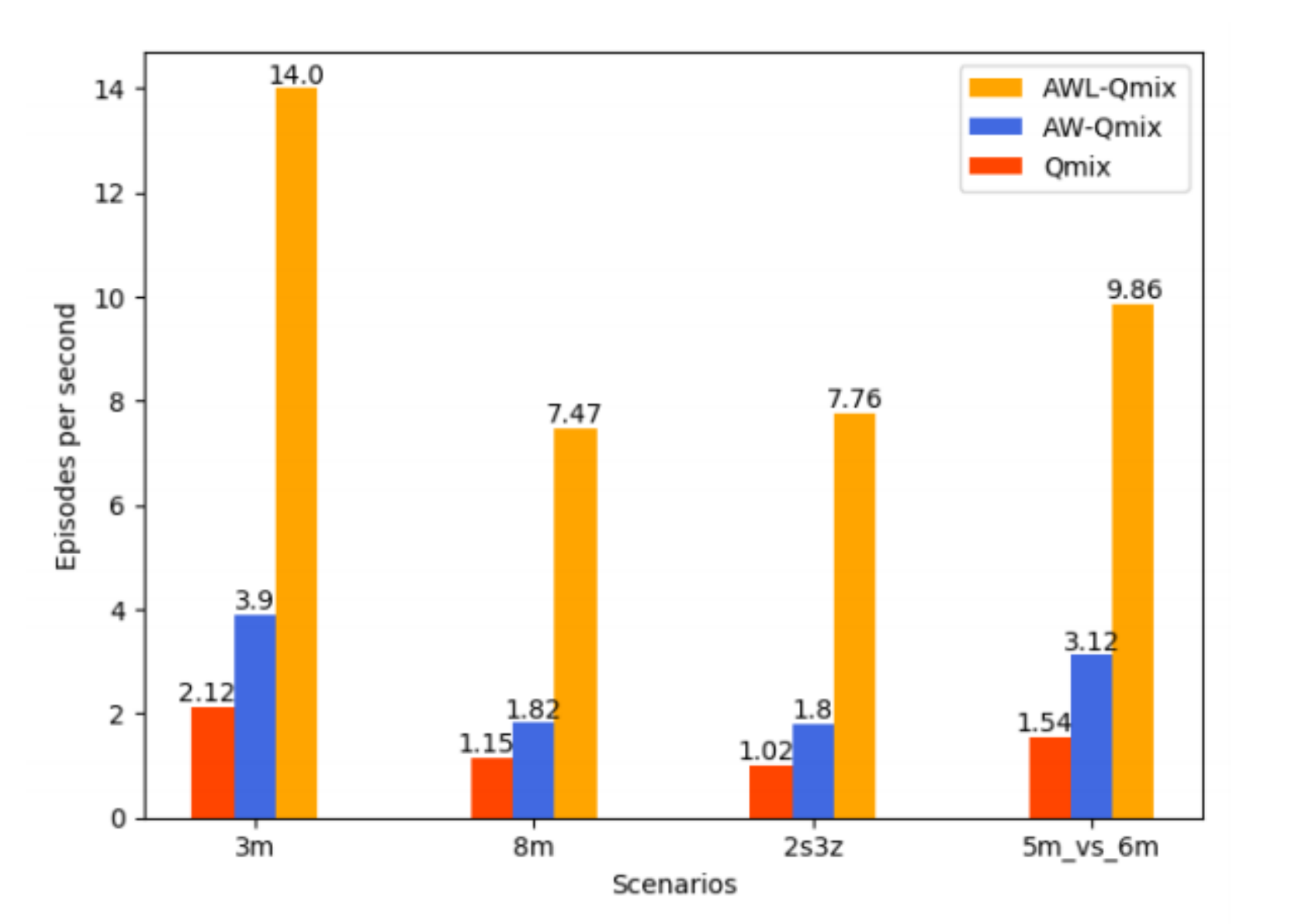}
    \caption{Sample collection speed in different scenarios}
    \label{fig:5}
\end{figure}

We use the Qmix, AW-Qmix, and AWL-Qmix to train in four scenarios, set the same collection number in sample for each method in the same scenario, record the sample collection time, and calculate the average collection speed of the sample.
The experimental results are shown in the Table \ref{tab:3} and Fig.~\ref{fig:5}.

\begin{table}[h]
    \centering
    \begin{tabular}{c|c|c|c}
    \toprule
    Scenario Name & Qmix & AW-Qmix & AWL-Qmix \\
    \hline
    $3m$ & 2.12 & 3.90 & \textbf{14.0}\\
    $8m$ & 1.15 & 1.82 & \textbf{7.47}\\
    $2s3z$ & 1.02 & 1.80 & \textbf{7.76}\\
    $5 m_{-} v s_{-} 6 m$ & 1.54 & 3.12 & \textbf{9.86}\\
    \bottomrule
    \end{tabular}
    \caption{Sample collection speed comparison.In four scenarios, AW-Qmix improves sample collection speed by almost 1.8, 1.6, 1.7, and 2.1 times over Qmix, whereas AWL-Qmix improves sample collection speed by nearly 6.6, 6.6, 7.6, and 2.1 times over Qmix.}
    \label{tab:3}
\end{table}

\begin{figure}[htbp]
    \centering
    \includegraphics[width=8cm]{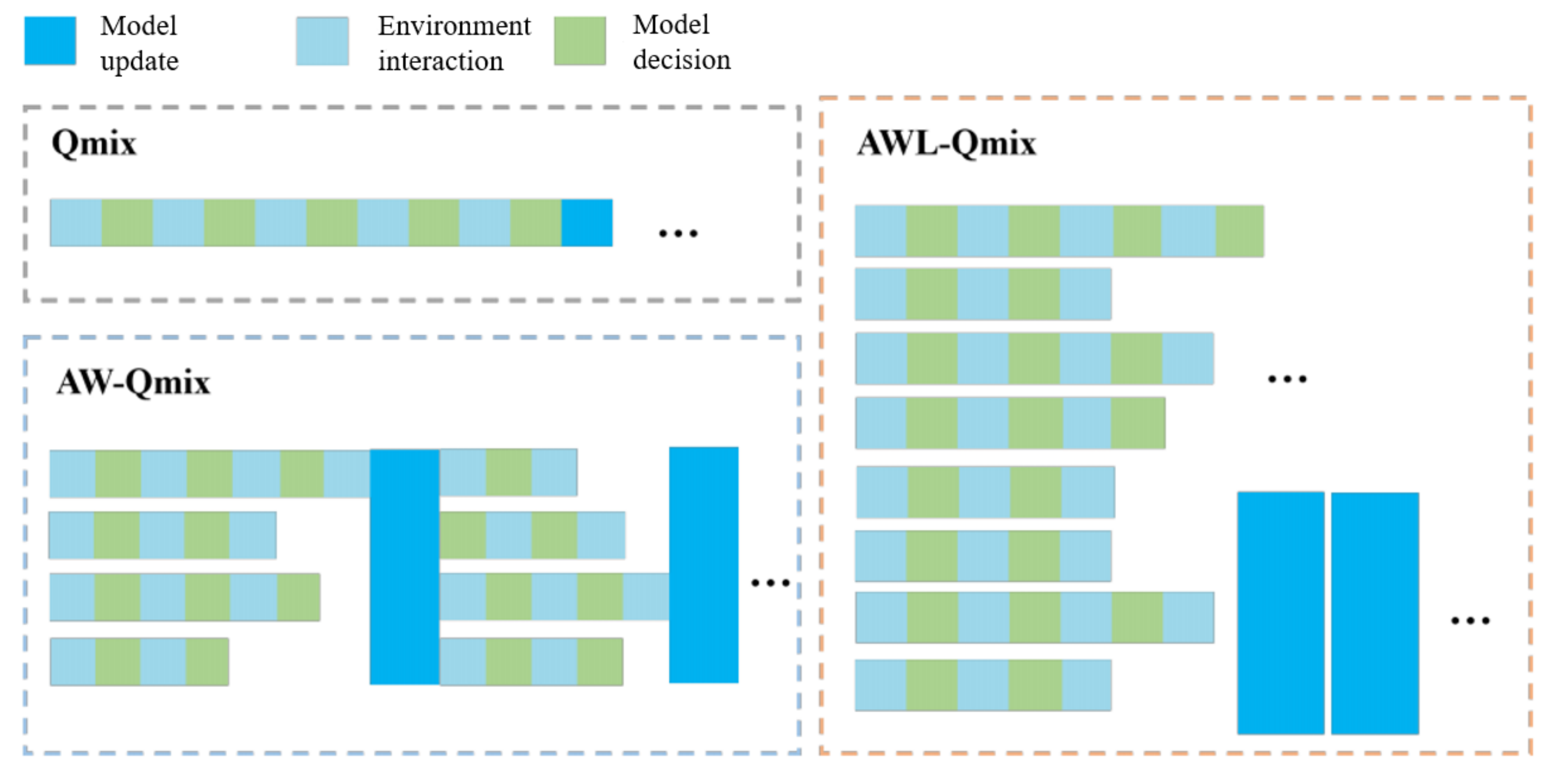}
    \caption{Iterative process of different models}
    \label{fig:6}
\end{figure}

The Qmix only manages one environment at a time. 
By opening numerous environments and leveraging asynchronous communication, the AW-Qmix allows for a minor increase in the update speed of model. 
Compared to the other two methods, the AWL-Qmix completely decouples model updating from environment interaction, allowing the model to be trained repeatedly using large-scale updated data which is shown in Fig.~\ref{fig:6}.

The counterfactual multi-agent policy gradient algorithm CoMA \cite{28} is used as a horizontal comparison approach in the experiment to further validate the performance of algorithm. 
In the tests, four actors are used for the AW-Qmix and four workers are used for the AWL-Qmix, with each worker including three actors. 
Depending on the difficulty, the interaction times of different environments are established for different scenarios. 
To ensure the scientificity of environment and rigor, the same amount of interactions are employed for each approach.

\subsubsection{SMAC}
\begin{figure}[h]
    \subfigure[]{
    \begin{minipage}[t]{0.5\linewidth}
    \includegraphics[width=4.3cm]{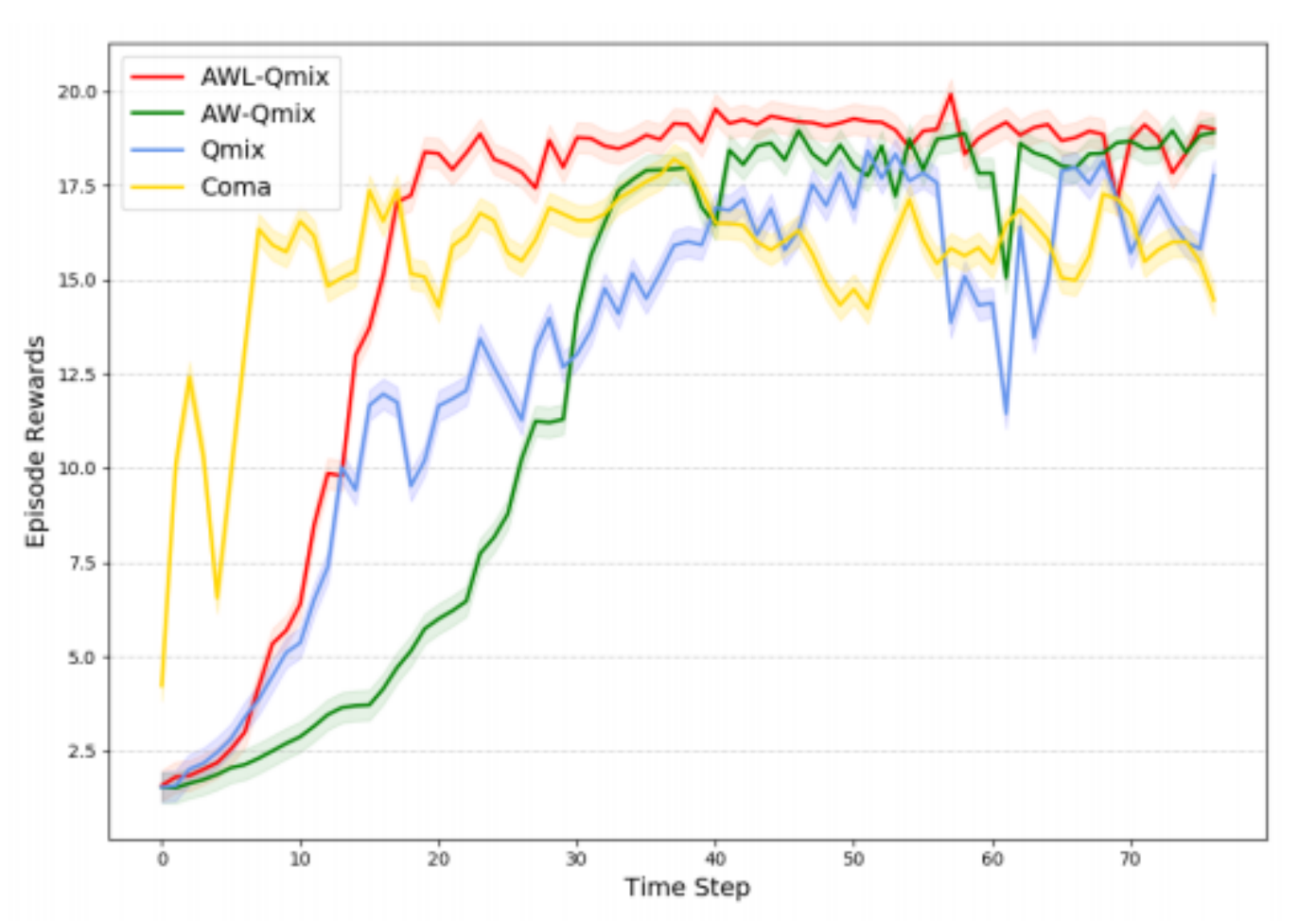}
    \label{subfig:3}
    \end{minipage}%
    }%
    \subfigure[]{
    \begin{minipage}[t]{0.5\linewidth}
    \includegraphics[width=4.3cm]{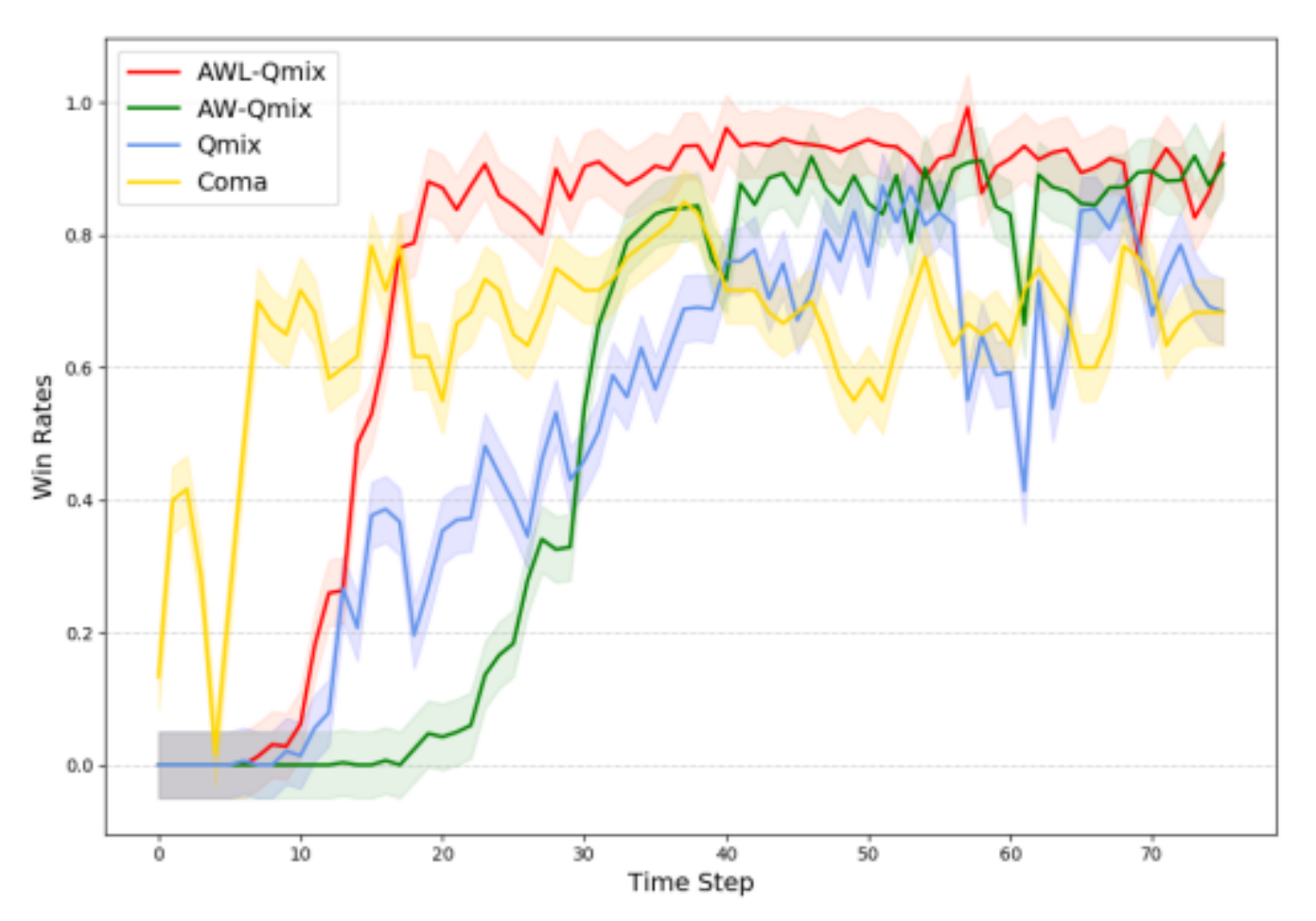}
    \label{subfig:4}
    \end{minipage}%
    }
    
    \subfigure[]{
    \begin{minipage}[t]{0.5\linewidth}
    \includegraphics[width=4.3cm]{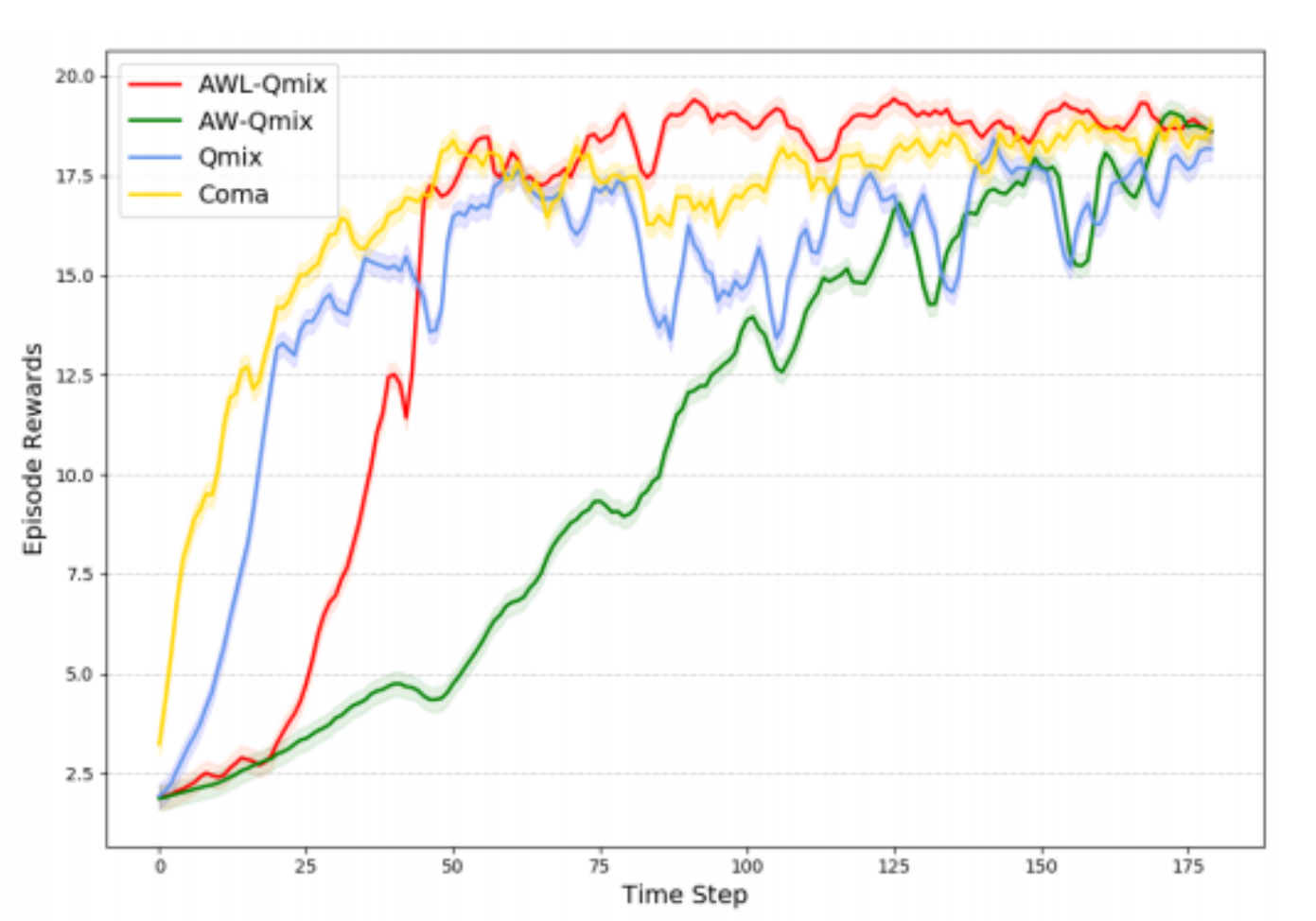}
    \label{subfig:5}
    \end{minipage}%
    }%
    \subfigure[]{
    \begin{minipage}[t]{0.5\linewidth}
    \includegraphics[width=4.3cm]{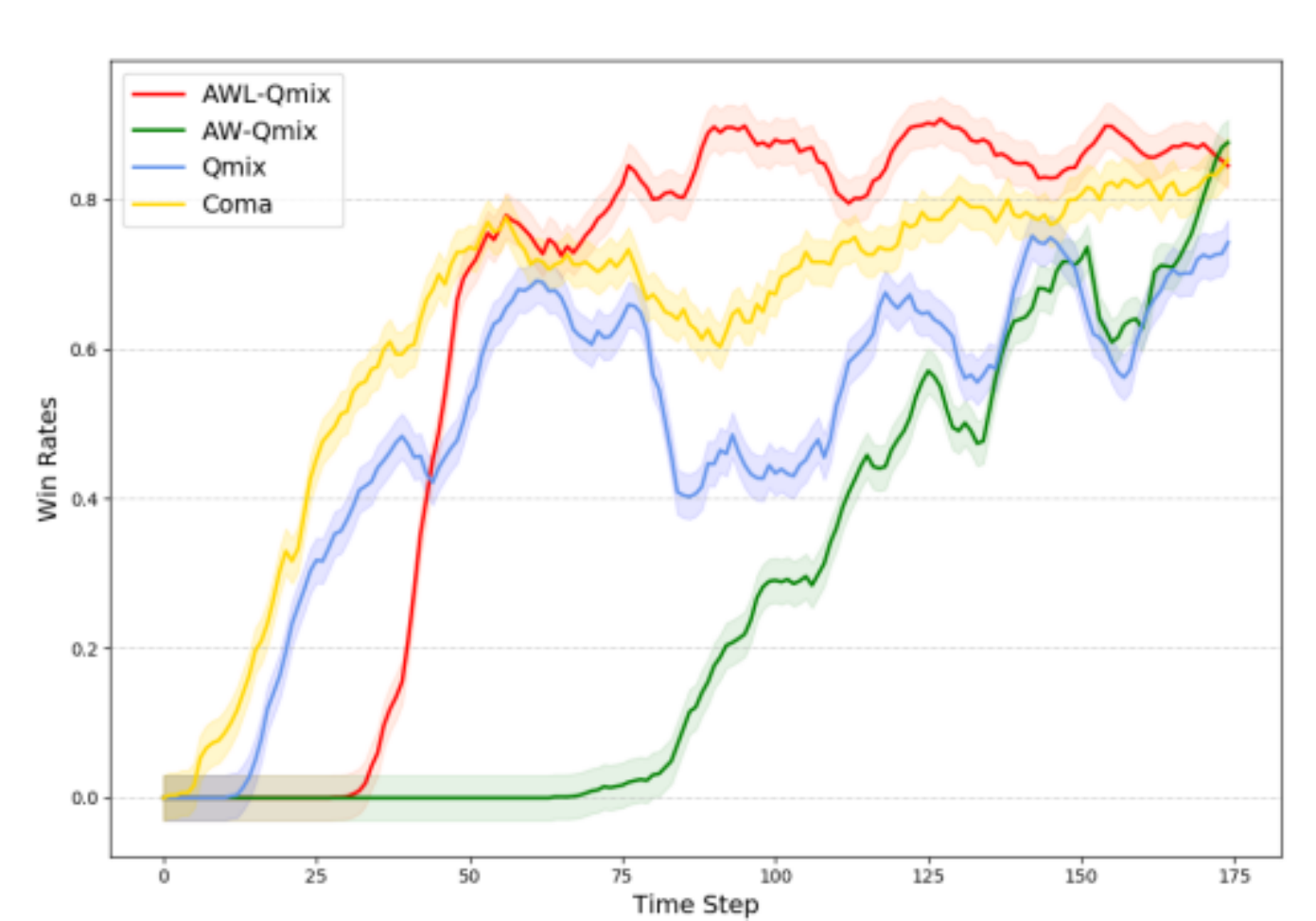}
    \label{subfig:6}
    \end{minipage}%
    }
    
    \subfigure[]{
    \begin{minipage}[t]{0.5\linewidth}
    \includegraphics[width=4.3cm]{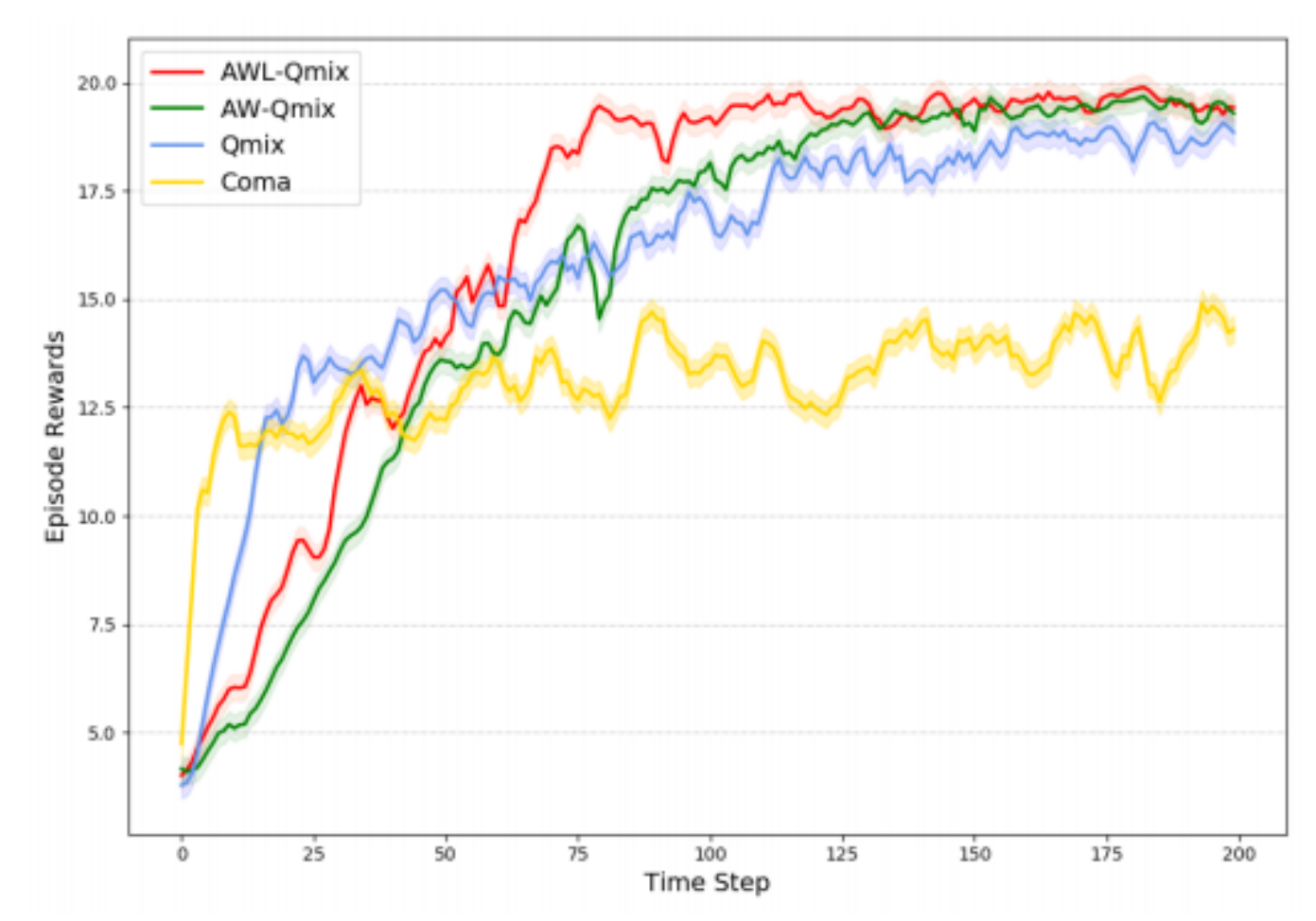}
    \label{subfig:7}
    \end{minipage}%
    }%
    \subfigure[]{
    \begin{minipage}[t]{0.5\linewidth}
    \includegraphics[width=4.3cm]{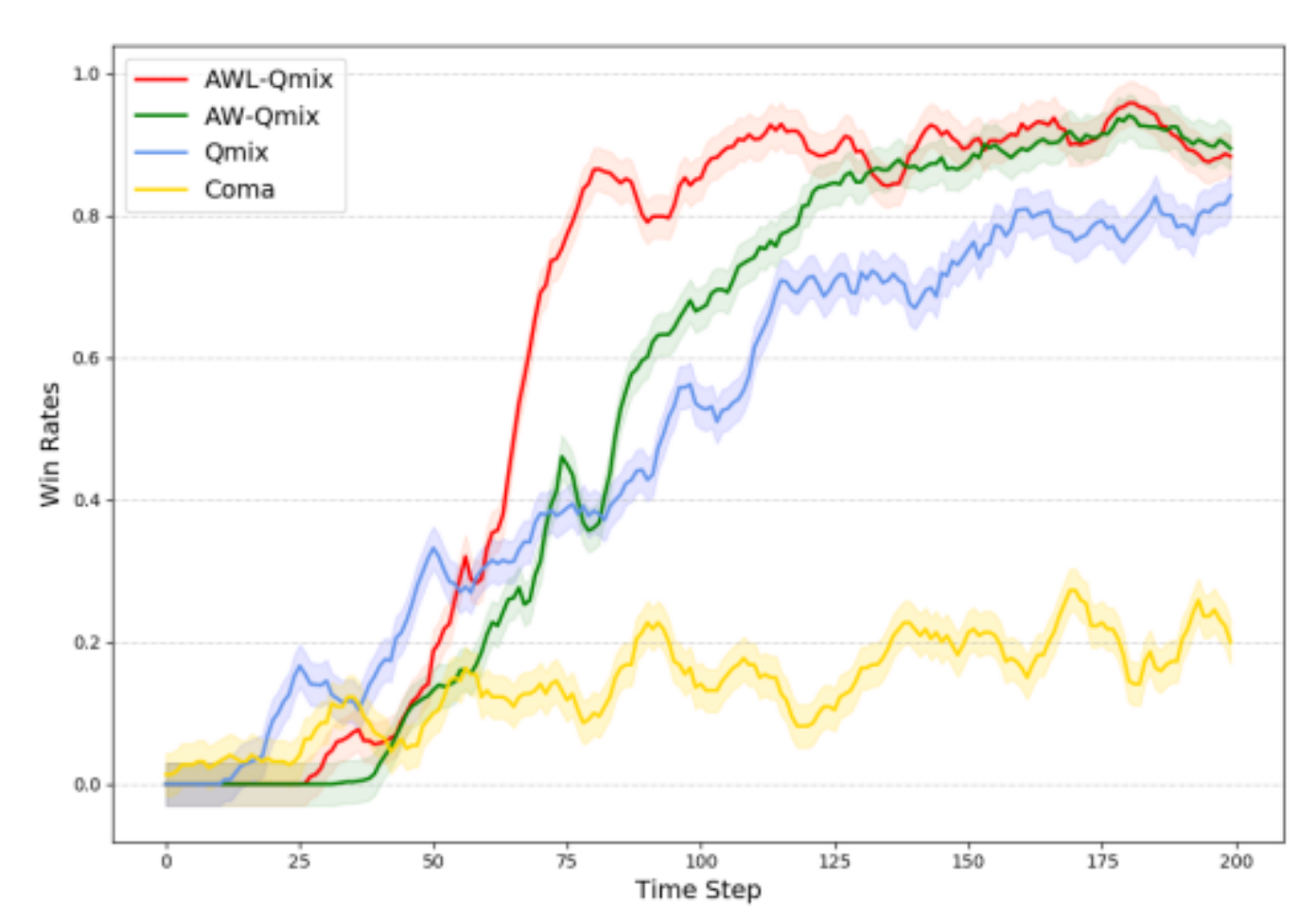}
    \label{subfig:8}
    \end{minipage}%
    }%
    
    \subfigure[]{
    \begin{minipage}[t]{0.5\linewidth}
    \includegraphics[width=4.3cm]{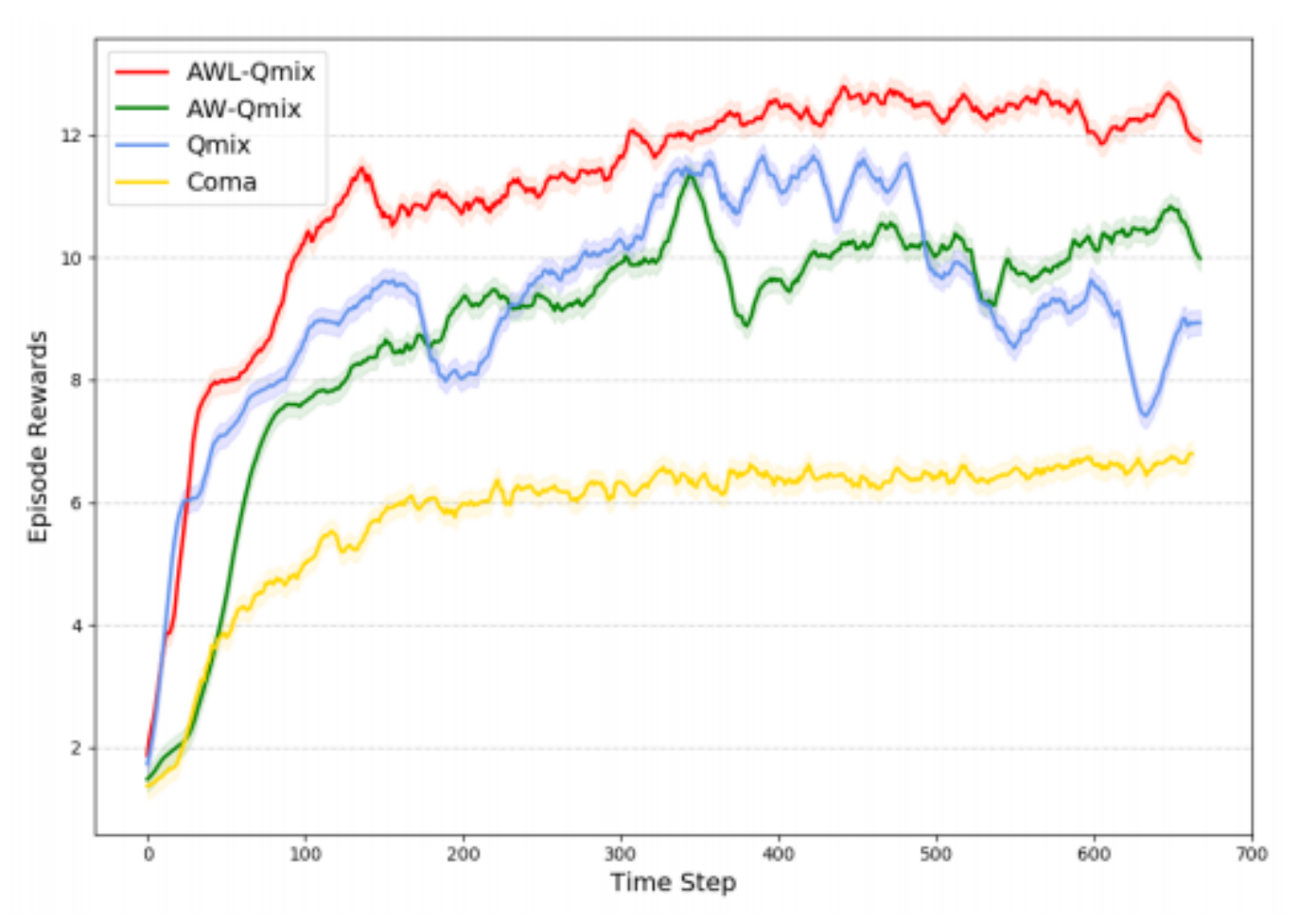}
    \label{subfig:9}
    \end{minipage}%
    }%
    \subfigure[]{
    \begin{minipage}[t]{0.5\linewidth}
    \includegraphics[width=4.3cm]{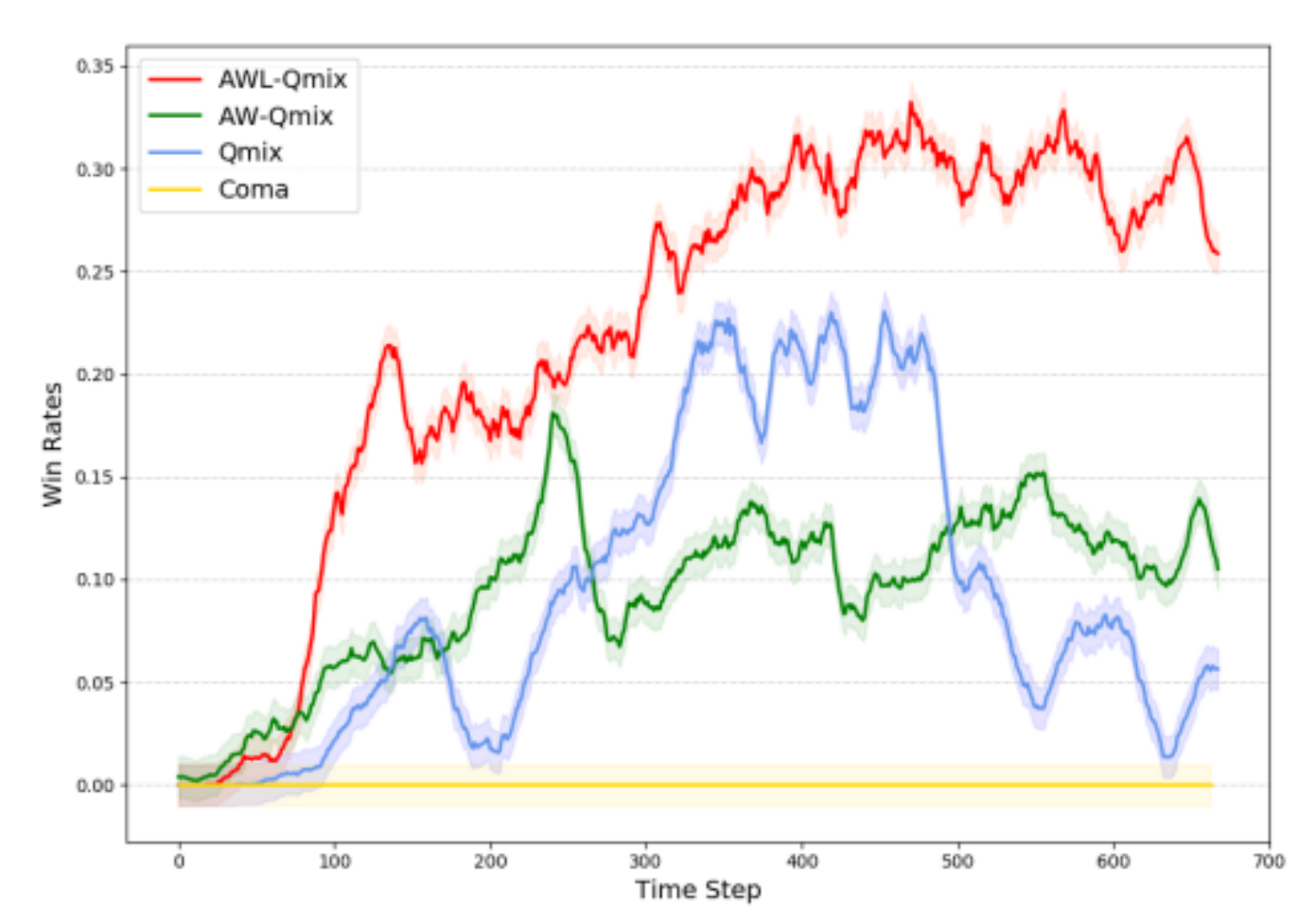}
    \label{subfig:10}
    \end{minipage}%
    }%
\caption{Algorithm performance comparison in four scenarios.Among them, \subref{subfig:3} and \subref{subfig:4} are the comparison between the reward value and the winning rate in the 3m scenario, \subref{subfig:5} and \subref{subfig:6} are in the 8m scenario, \subref{subfig:7} and \subref{subfig:8} are in the 2s3z scenario, and \subref{subfig:9} and \subref{subfig:10} are in the $5 m_{-} v s_{-} 6 m$ scenario. \subref{subfig:3}, \subref{subfig:5}, \subref{subfig:7}, \subref{subfig:9} are the comparison of reward values in the four scenarios, and \subref{subfig:4}, \subref{subfig:6}, \subref{subfig:8}, \subref{subfig:10} are the comparison of the winning rates in the four scenarios.}
\label{fig:7}
\end{figure}

It can be seen from Fig.~\ref{subfig:3}, \ref{subfig:4}, \ref{subfig:5},  \ref{subfig:6} that within the same time step in the 3m scenario, AW-Qmix and AWL-Qmix converge to the optimal value earlier. 
At this time, Qmix is still fluctuating and does not reach a stable reward value. Although CoMA tends to converge earlier, the reward value achieved is low and the performance is at a disadvantage. 

In the 8m scenario, as the number of agents increases, the convergence of AW-Qmix is hindered, but AWL-Qmix still maintains better performance and stability.
The possible reason is that AW-Qmix adopts a single-model decision-making network structure, and samples and updates after collecting samples in each environment. 
When the model encounters conflicting samples, the information brought by the samples may cause the model to converge in different directions, resulting in the model needing more exploration to find the correct convergence direction. 
The multi-model network and shared model technology adopted by AWL-Qmix can allow multiple decision networks to be updated in time, and the update directions of each decision network are consistent, thereby reducing the possibility of conflict between interactive samples.

It is shown that AW-Qmix and AWL-Qmix are marginally better than Qmix in terms of convergence speed and final performance, and are much better than CoMA in the heterogeneous symmetric scenario 2s3z in Fig.~\ref{subfig:7}, \ref{subfig:8}.
The most demanding and complex case for the agent is the symmetric $5 m_{-} v s_{-} 6 m$ scenario. 
It necessitates the learning of the agent from a disadvantageous situation to defeat the enemy. 
CoMA is tough to defeat the opponent in this difficult scenario, as shown in Fig.~\ref{subfig:9}, \ref{subfig:10}, and AWL-Qmix has a significant advantage in rewards and winning rate curves over the other three approaches.

\subsubsection{MaCA}
We install distributed decoupling architectures of various scales on the experimental platform MaCA \cite{29} to further test the generality of the proposed DMCAF on diverse platforms.

\begin{figure}[h]
    \centering
    \includegraphics[width=8cm]{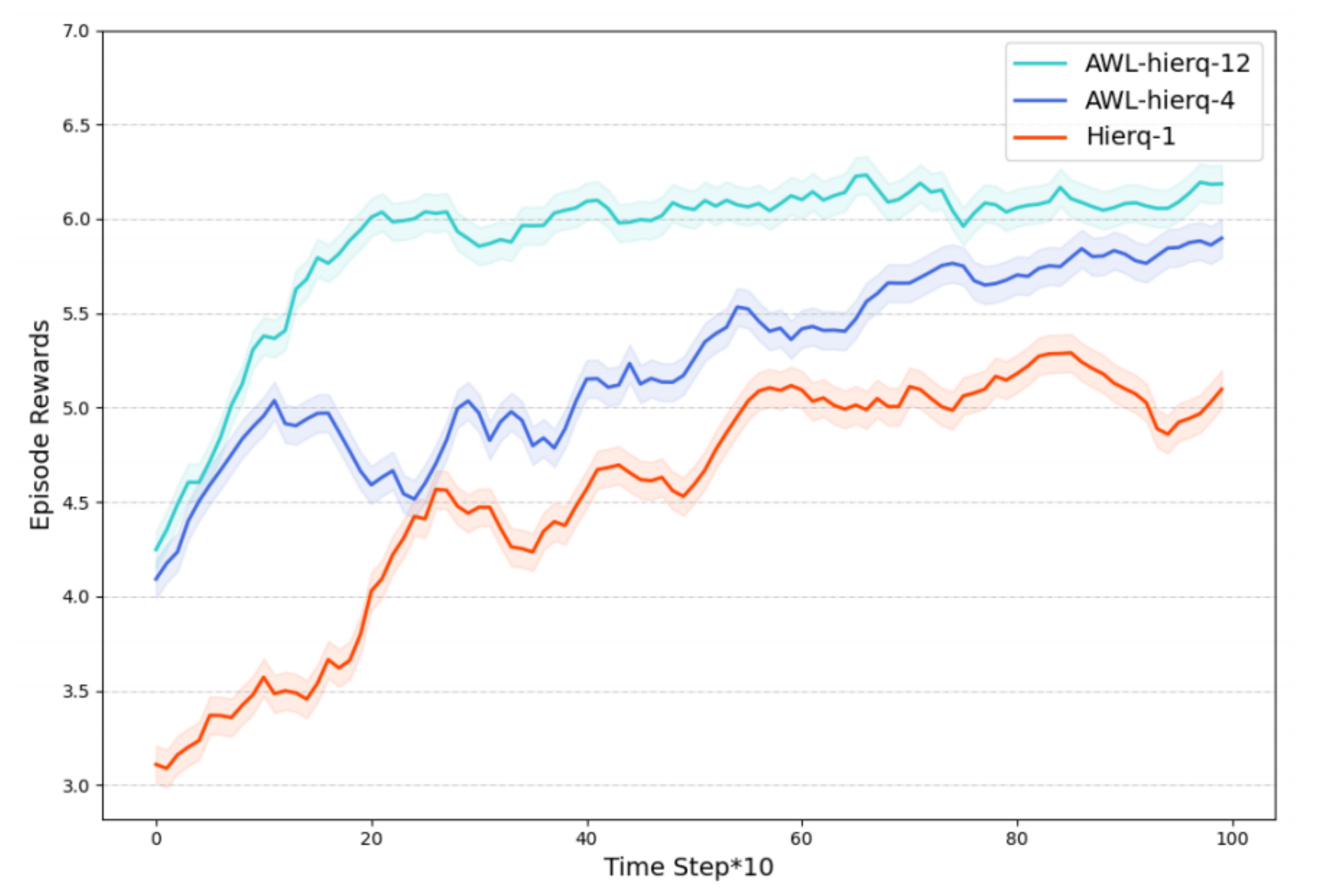}
    \caption{Performance Comparison in MaCA.The Hierq-1 approach is an upgraded hierarchical Qmix algorithm. The AWL-hierq-4 approach deploys two workers, each with two actors, resulting in four interactive environments, and the AWL-hierq-12 approach deploys three workers, each with four actors, resulting in a total of twelve interactive environments.}
    \label{fig:8}
\end{figure}

The performance comparison of the three methods is shown in Fig.~\ref{fig:8}. 
It can be seen that AWL-hierq-12 and AWL-hierq-4 have a certain advantage in reward score compared with Hierq-1, and this advantage increases with the increase of the distributed scale. 
This result once again verifies the effectiveness of DMCAF, and also demonstrates the application of DMCAF in large-scale reinforcement learning training.

\section{Conclusion}
In this paper, we introduced two distributed frameworks for MARL.
In the actor-worker framework, the idea of distributed training in the field of single-agent reinforcement learning is extended to the field of MARL.
And a multi-agent asynchronous communication algorithm based on actor-worker is designed to solve the urgent needs of sample collection of MARL. Then, we extended the actor-worker to the actor-worker-learner framework. In this framework, the environment interaction is decoupled from the model iteration process, which further increases the diversity of training sample. Finally, the effectiveness of our method is investigated on the StarCraft II reinforcement learning benchmark platform and the MaCA platform. In the future work, we will try to introduce a multi-agent-oriented priority experience replay mechanism based on the distributed algorithm for optimization.

\bibliographystyle{ieeetr}
\bibliography{paper}
\end{document}